\documentclass{article}
\usepackage[final,nonatbib]{neurips_2023}

\usepackage[utf8]{inputenc} %
\usepackage[T1]{fontenc}    %
\usepackage[pdfborder={0 0 0}]{hyperref}       %
\usepackage{url}            %
\usepackage{booktabs}       %
\usepackage{amsfonts}       %
\usepackage{nicefrac}       %
\usepackage{microtype}      %
\usepackage{xcolor}         %

\newif\ifpreface
\prefacetrue %

\newif\ifabstract
\abstracttrue %

\newif\ifcontent
\contenttrue %
\makeatletter
\renewcommand{\@noticestring}{Preprint version, see \url{https://doi.org/10.1515/9783110785944-005}}
\makeatother

\input{preamble.tex}

\title{Data Aggregation for Hierarchical Clustering}

\author{%
  Erich Schubert\\
  TU Dortmund University\\
  \texttt{erich.schubert@tu-dortmund.de} \\
  \And
  Andreas Lang\\
  TU Dortmund University\\
  \texttt{andreas.lang@tu-dortmund.de} \\
}

\begin{document}
\maketitle

\bgroup
\let\subsubsection\subsection
\let\subsection\section
\let\tsum\undefined\newcommand{\tsum}{\textstyle\sum}
\let\CF\undefined\newcommand{\CF}{\mathrm{CF}}
\let\SSE\undefined\newcommand{\SSE}{\mathrm{SSE}}
\let\Var\undefined\newcommand{\Var}{\operatorname{Var}}
\let\norm\undefined\newcommand{\norm}[1]{\left\lVert#1\right\rVert}
\let\snorm\undefined\newcommand{\snorm}[1]{\left\lVert\smash{#1}\mathstrut\right\rVert}
\let\AuB\undefined\newcommand{\AuB}{A{\cup}B}
\let\nAB\undefined\newcommand{\nAB}{n_{\!A\!B}}
\let\nAC\undefined\newcommand{\nAC}{n_{\!A\!C}}
\let\nBC\undefined\newcommand{\nBC}{n_{\!B\!C}}
\let\nABC\undefined\newcommand{\nABC}{n_{\!A\!B\!C}}
\let\nABSq\undefined\newcommand{\nABSq}{{\nAB}^2}
\let\nABCSq\undefined\newcommand{\nABCSq}{{\nABC}^2}
\let\O\undefined\newcommand{\O}[1]{\ensuremath{{O\mathopen{}(#1)\mathclose{}}}}

\ifabstract
\begin{abstract}%
Hierarchical Agglomerative Clustering (HAC) is likely the earliest and most flexible clustering method,
because it can be used with many distances, similarities, and various linkage strategies.
It is often used when the number of clusters the data set forms is unknown and some sort of hierarchy in the data is plausible.
Most algorithms for HAC operate on a full distance matrix, and therefore require quadratic memory.
The standard algorithm also has cubic runtime to produce a full hierarchy.
Both memory and runtime are especially problematic in the context of embedded or otherwise very resource-constrained systems.
In this section, we present how data aggregation with BETULA,
a numerically stable version of the well known BIRCH data aggregation algorithm,
can be used to make HAC viable on systems with constrained resources with only small losses on clustering quality,
and hence allow exploratory data analysis of very large data sets.
    
\end{abstract}
\fi

\textcolor{red}{This is a preprint of Erich Schubert and Andreas Lang. ``5.3 Data Aggregation for Hierarchical Clustering''.
In: Machine Learning under Resource Constraints -- Fundamentals (Volume 1) edited by Katharina Morik and Peter Marwedel, Berlin, Boston: De Gruyter, 2023, pp.{} 215-226.
\url{https://doi.org/10.1515/9783110785944-005}}

\ifcontent

\subsection{Introduction}
Hierarchical Agglomerative Clustering (HAC) \index{Hierarchical clustering}
is a popular clustering method that is especially useful if a hierarchy of clusters exists in the data set.
Initially, each data entry is seen as a cluster of one.
In each hierarchy level the two clusters with the least distance (c.f. Section~\ref{sec:lang_linkage}) between them are combined until the whole data set is in one cluster.
Another commonly used name, Simple Agglomerative Hierarchical Nesting (SAHN), reflects this easy-to-understand core idea.
The standard algorithm used for HAC, known as AGNES~\cite{Kaufman/Rousseeuw/90a},\index{AGNES clustering}
requires the pairwise distances between all data points to be stored in a distance matrix,
and when merging clusters, two columns and rows in this matrix are combined using the Lance-Williams equations \cite{Lance/Williams/66a,Lance/Williams/67a}.
AGNES can be utilized with different primary distance functions, but also with different cluster distances (commonly called linkages), see Section~\ref{sec:lang_linkage}.
Hierarchical Agglomerative Clustering, like many other clustering methods, is a rather resource-hungry process commonly implemented using $\O{N^2}$ memory and $\O{N^2}$ to $\O{N^3}$ time, depending on the exact algorithm implemented.
One possibility to reduce the resource demands for big data or when using small embedded systems is data aggregation.
The BIRCH (Balanced Iterative Reducing and Clustering using Hierarchies) \cite{Zhang/etal/96a,Zhang/etal/97a}\index{BIRCH clustering}
algorithm is a well-known data aggregation technique for clustering.
BIRCH is a multi-step clustering algorithm that aggregates the data into a tree structure known as CF-tree before the actual clustering.
We will first review some fundamentals of hierarchical clustering, and
then discuss an improved version of BIRCH, called BETULA \cite{Lang/Schubert/2020a,Lang/Schubert/2021a},
that avoids some numerical problems in the original BIRCH.
We then show how it can be used to accelerate HAC for big data, and reduce its memory requirements. 

\subsection{Hierarchical Clustering Linkages} 
\label{sec:lang_linkage}

Because Hierarchical Agglomerative Clustering is based on the idea of always merging the two closest clusters,
we need to define a suitable distance of clusters, not just of single points.
Usually, we want this distance to be consistent with our distance of single points.
This notion of ``cluster distance'' is commonly called the ``linkage'' criterion.
The choice of linkages affect greatly how the resulting clusters look,
but they also influence which algorithms can be used.

The two most widely known linkage strategies are single-link and complete-link,
where the distance of two clusters is defined as the minimum respectively
maximum distance of any two points. However, many other linkages have been
proposed in literature, many already back in the 1950s by, e.g.,
\citet{Sneath/57a,McQuitty/57a,Sokal/Sneath/63a,Wishart/69a}.
More recent proposals include Mini-Max \cite{Ao/etal/2005a} and Medoid linkages \cite{Herr/etal/2016a,Miyamoto/etal/2016a,Schubert/2021a}.
Several (but not all) linkages can be expressed in terms of Lance-Williams recurrences \cite{Lance/Williams/66a,Lance/Williams/67a},
which offer computational advantages.
WPGMA (McQuitty) and WPGMC (Median linkage),
can only be defined in terms of a recurrence, and do not have a closed-form
based only on the sets of points.
The Lance-Williams formula is as follows:
\begin{align}
d(A {\cup} B, C) =& \alpha_A d(A,C) + \alpha_B d(B,C) + \beta d(A,B)
+ \gamma|d(A,C){-}d(B,C)|
\label{eq:LanceWilliams}
\;.
\end{align}
Different linkage strategies can be defined in terms of the factors
$\alpha_A$, $\alpha_B$, $\beta$, and~$\gamma$ as given in Tab.~\ref{tab:linkages}. 
These may depend on the sizes of the clusters $A$, $B$, and~$C$,
which we denote as $n_A$, $n_B$, and~$n_C$.
For brevity, we use the shorthand $\nAB:=n_{A{\cup}B}=n_A{+}n_B$,
and $\nABC:=n_{A{\cup}B{\cup}C}=n_A{+}n_B{+}n_C$.
An additional -- and often overlooked -- detail is the initialization of the distance matrix.
While single, complete, and group-average linkage work with any distance,
\index{Single linkage clustering}\index{Complete linkage clustering}%
the centroid, Ward and median methods need to be initialized with squared distances,
\index{Centroid linkage clustering}\index{Ward linkage clustering}\index{Median linkage clustering}%
and are closely tied to the Euclidean distance and variance.
Ignoring this initialization difference (and interpretation of the output)
can easily yield to receiving incorrect results \cite{Murtagh/Legendre/2014a}.
The reason becomes apparent when considering the objective function of
the clustering, respectively the closed-form, e.g., Eqs.~\eqref{eq:avglinkage} to \eqref{eq:ward}.

\begin{table}\centering%
\caption{Common linkages in terms of Lance-Williams factors}
\label{tab:linkages}
\renewcommand{\arraystretch}{1.2}%
\begin{tabular}{lccccl}
\toprule
Linkage & $\alpha_A$ & $\alpha_B$ & $\beta$ & $\gamma$ & Init. \\
\midrule
Single & $1/2$ & $1/2$ & $0$ & $-1/2$ & $d(a,b)$ 
\\[.2ex]
Complete & $1/2$ & $1/2$ & $0$ & $1/2$ & $d(a,b)$ 
\\[.8ex]
Group-average (UPGMA) & $\displaystyle\frac{n_A}{\nAB}$ & $\displaystyle\frac{n_B}{\nAB}$ & $0$ & $0$ & $d(a,b)$ 
\\[1.2ex]
McQuitty (WPGMA) & $1/2$ & $1/2$ & $0$ & $0$ & $d(a,b)$ 
\\[.8ex]
Centroid (UPGMC) & $\displaystyle\frac{n_A}{\nAB}$ & $\displaystyle\frac{n_B}{\nAB}$ & $-\displaystyle\frac{n_A\cdot n_B}{\nABSq}$  & $0$ & $d(a,b)^2$ 
\\[1.2ex]
Median (WPGMC) & $1/2$ & $1/2$ & $-1/4$ & $0$ & $d(a,b)^2$ 
\\[.8ex]
Ward  & $\displaystyle\frac{\nAC}{\nABC}$ & $\displaystyle\frac{\nBC}{\nABC}$ & $-\displaystyle\frac{n_C}{\nABC}$ & $0$ & $d(a,b)^2$
\\[.8ex]
\bottomrule
\end{tabular}
\end{table}

Consider single-linkage first (and, by substituting $\max$ for $\min$,
complete-linkage). Here the aim is to merge clusters $A$ and $B$ with the
smallest distance between their points, i.e., with the smallest
$d_{\text{single}}(A,C):=\min_{a\in A,c\in C} d(a,c)$.
If both clusters consists of a single element, we obviously have
$d_{\text{single}}(\{a\},\{c\})=d(a,c)$, and we can recursively compute this linkage using
$d_{\text{single}}(A\cup B,C) = \min\{d_{\text{single}}(A,C),d_{\text{single}}(B,C)\}$.
It is easy to see that the weights given in Tab.~\ref{tab:linkages} correspond to using the minimum respective maximum.

Group-average linkage, also known as 
Unweighted Pair Group Method with Arithmetic mean (UPGMA),
is another very intuitive linkage, %
and is often considered one of the best to use in practice.
The idea is to capture the average distance between elements
from different clusters, i.e.,
$d_{\text{avg}}(A,C):=\tfrac{1}{n_A n_C} \sum_{a\in A} \sum_{c\in C}
d(a,c)$. Clearly, for one-elemental clusters,
we have $d_{\text{avg}}(\{a\},\{c\})=d(a,c)$. The recursive
computation formula is easy to derive:
\begin{align}
d_{\text{avg}}(A\cup B,C) =&
\tfrac{1}{\nAB n_C} \bigg( \sum_{a\in A} \sum_{c\in C} d(a,c)
+ \sum_{b\in B} \sum_{c\in C} d(b,c) \bigg)
\notag
\\
=& \tfrac{1}{\nAB n_C} \left( n_A n_C d_{\text{avg}}(A,C)
+ n_B n_C d_{\text{avg}}(B,C) \right)
\notag
\\
=& \tfrac{n_A}{\nAB} d_{\text{avg}}(A,C) + \tfrac{n_B}{\nAB} d_{\text{avg}}(B,C)
\enskip. %
\label{eq:avglinkage}
\end{align}
The literature terminology \enquote{weighted} (going back to
\citeauthor{Sokal/Sneath/63a} \cite{Sneath/57a,Sokal/Sneath/63a}) can be confusing:
it refers to the influence each point has.
In \enquote{unweighted} group average, each \emph{object} has the same weight
(and, hence, the weight of each cluster is proportional the number of objects
contained),
whereas in the \enquote{weighted} version i.e McQuitty and Median linkage, each \emph{cluster} has the same
weight (and, hence, each object in a larger cluster has a reduced weight).
As easily seen in Tab.~\ref{tab:linkages}, both ``weighted'' versions correspond
to their ``unweighted'' counterparts if we fix the cluster sizes to a constant $n_A=n_B:=1$,
i.e., ignoring the cluster sizes when merging.

McQuitty's Weighted Pair-Group Method with Arithmetic mean (WPGMA~\cite{McQuitty/57a})
\index{McQuitty's Weighted Pair-Group Method with Arithmetic mean (WPGMA)}%
can be recursively defined as
$d_{\text{McQ}}(A\cup B,C) =
\tfrac12 (d_{\text{McQ}}(A,C) + d_{\text{McQ}}(B,C))$,
which introduces an unfortunate dependency on the \enquote{merge history}
of the child clusters $A$ and $B$. Given three objects $a,b,c$, merging
$a$ and $b$ first, then with $c$ may yield a different result than first
merging one of the other pairs.
A similar argument holds for median linkage (WPGMC), discussed below.

Unweighted Pair-Group Method using Centroids (UPGMC), also known as
centroid linkage, combines clusters by the distance of the cluster means $\mu_A = \tfrac{1}{|A|}\sum_{x \in A}x$,
i.e., always merges the smallest $d_{\text{cent}}(A,C)=\norm{\mu_A-\mu_C}$.
Computing the distances between the means explicitly requires many
additional distance computations, and hence is slower and less resource-efficient than a recurrent approach. %
But there is a special relationship between the mean, the variance,
and squared Euclidean distance that we can exploit to compute this
special case elegantly with a recurrence.
We discuss this relationship, without loss of generality, only for univariate data,
because squared Euclidean is simply the sum of the squared variates.
We then have $\norm{\mu_A{-}\mu_C}^2=\mu_A^2+\mu_C^2-2\mu_A\mu_C$ and obtain
\begin{align}
d_{\text{cent}}(\AuB,C)=& \tfrac{n_A}{\nAB} d_{\text{cent}}(A,C)
+ \tfrac{n_B}{\nAB} d_{\text{cent}}(B,C)
- \tfrac{n_An_B}{\nABSq} d_{\text{cent}}(A,B)
\notag
\\
=&\mu_{AB}^2+\mu_C^2-2\mu_{AB}\mu_C
=\norm{\mu_{AB} - \mu_C}^2
\enskip.
\label{eq:centroidlinkage}
\end{align}

This means that for squared Euclidean distances, we can compute the distance
of the means without computing the means themselves. Hence, we need to initialize
the distance matrix with squared Euclidean distances, and also need to interpret
the resulting linkage distances as such squared values.

The idea of median linkage (or Weighted Pair-Group Method using Centroids, WPGMC)
is to minimize the distance of the medians,
$\norm{m_{A\cup B}-m_C}$, where the median is recursively defined as
$m_{A\cup B} = \tfrac12(m_A + m_B)$, the midpoint of the previous medians.
For squared Euclidean distances, we again have
a recurrent formula; the derivation is exactly as for centroid
linkage, but with fixed $n_A=n_B=1$.
Median linkage and centroid linkage have the oddity that the distance
$d(A\cup B,C)$ can be less than the distance of $d(A,C)$, which can yield
non-monotone dendrograms: if we draw a tree representing the cluster merges,
and use the linkage distance as height of a branch,
the resulting tree does not monotonously grow.
Such anomalies in the trees also referred to as inversions, and can only
if a linkages does not have the reducibility property of \citet{Bruynooghe/77a}).
Intuitively, this happens when the new center is between two well-separated
clusters, and then closer to a third than either of the two, as illustrated
in Fig.~\ref{lang:non-monotone}. This can cause
undesirable results, and these linkages should be used with care.

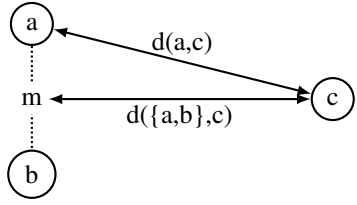
\begin{figure}\centering
\begin{tikzpicture}
\node[draw, thick, circle] (a) at (0,+1) {a};
\node[draw, thick, circle] (b) at (0,-1) {b};
\draw[thick, densely dotted] (a) -- (b);
\node[thick, circle, fill=white, inner sep=1pt] (m) at (0,0) {m};
\node[draw, thick, circle] (c) at (4,0) {c};
\draw[thick,<->,>=latex] (a) -- (c) node[midway, above, inner sep=2pt] {d(a,c)};
\draw[thick,<->,>=latex] (m) -- (c) node[midway, below, inner sep=2pt] {d(\{a,b\},c)};
\end{tikzpicture}
\caption{Example why Median and Centroid linkages are non-monotone: the midpoint $m$ of the merged cluster $\{a,b\}$
is closer to $c$ than any of its clusters members $a$ and $b$ were.}
\label{lang:non-monotone}
\end{figure}

The popular Ward linkage optimizes the criterion \cite{Wishart/69a,Anderberg/73a,Kaufman/Rousseeuw/90a}:
\begin{align}
d_{\text{Ward}}(A,B) =& \tfrac{2n_A\cdot n_B}{\nAB}\snorm{\mu_A-\mu_B}^2
\enskip. %
\label{eq:ward}
\end{align}
The factor 2 in this equation ensures that $d_{\text{Ward}}(\{a\},\{b\})=\snorm{a,b}^2$,
as desired for one-elemental clusters.
This criterion can be described as
\enquote{minimum increase in the sum of squares}~\cite{Podani/89a},
which may come as a surprise given that the equation only uses the means,
and does not appear to contain the sum of squares.
The reader may have noticed that $k$-means clustering also
minimizes the sum of squares. The main difference here is that Ward
linkage imposes a hierarchical structure on the result, whereas $k$-means
imposes a flat partitioning into $k$ partitions. Usually, the result of
Ward linkage cut at $k$ partitions will be (often substantially) worse
than that of $k$-means (for the consistency reasons explained in \citet{Schubert/2021a}
for the case of medoid linkage),
but on the other hand, $k$-means results for varying
$k$ will usually not nest into a hierarchy of clusters.
Equation~\refeq{eq:ward} can be obtained from rewriting the increase
in the sum of squares via the König-Huygens theorem:
\begin{align*}
d_{\text{Ward}}(A,B)
=&
 \tsum\limits_{x\in \AuB} \norm{x{-}\mu_{AB}}^2
 - \tsum\limits_{a\in A} \norm{a{-}\mu_A}^2
 - \tsum\limits_{b\in B} \norm{b{-}\mu_B}^2
 \\=&
 \tfrac{2n_An_B}{\nAB} \norm{\mu_A{-}\mu_B}^2 
 \end{align*}
The Lance-Williams recurrence given in Tab.~\ref{tab:linkages} follows
(full derivation omitted):
\begin{align*}
d_{\text{Ward}}(\AuB,C)
=&
 \tfrac{2\nAB n_C}{\nABC} \norm{\mu_{AB}{-}\mu_C}^2
 = %
 \tfrac{2\nAB n_C}{\nABC}
 \snorm{\tfrac{n_A}{\nAB}\mu_A{+}\tfrac{n_B}{\nAB}\mu_B{-}\mu_C}^2
\\=&
\tfrac{\nAC}{\nABC}d_{\text{Ward}}(A,C)
+ \tfrac{\nBC}{\nABC}d_{\text{Ward}}(B,C) %
- \tfrac{n_C}{\nABC}d_{\text{Ward}}(A,B)
\end{align*}

\subsection{The Cluster Feature Tree (CF-Tree)}

\begin{figure}\centering
\begin{tikzpicture}[scale=.7]
\tikzstyle{cf}=[draw, anchor=center, inner sep=2pt, minimum width=.7cm, minimum height=.35cm, font=\small\strut]

\draw[thick] (-.7,.45) rectangle +(4.4,-.9);
\node[cf] at (0,0) {$\CF_1$};
\node[cf] at (1,0) {$\CF_2$};
\node[cf] at (2,0) {$\CF_3$};
\node[cf] at (3,0) {\ldots};

\draw[->,>=latex,thick] (0,-.3) -- (-2.5,-.8);
\draw[thick] (-4.7,-.85) rectangle +(4.4,-.9);
\node[cf] at (-4,-1.3) {$\CF_1$};
\node[cf] at (-3,-1.3) {$\CF_2$};
\node[cf] at (-2,-1.3) {$\CF_3$};
\node[cf] at (-1,-1.3) {\ldots};

\draw[->,>=latex,thick] (1,-.3) -- (2.5,-.8);
\draw[thick] (.3,-.85) rectangle +(4.4,-.9);
\node[cf] at (1,-1.3) {$\CF_1$};
\node[cf] at (2,-1.3) {$\CF_2$};
\node[cf] at (3,-1.3) {$\CF_3$};
\node[cf] at (4,-1.3) {\ldots};

\draw[->,>=latex,thick] (2,-.3) -- (7,-.8);
\draw[thick] (5.3,-.85) rectangle +(3.4,-.9);
\node[cf] at (6,-1.3) {$\CF_1$};
\node[cf] at (7,-1.3) {$\CF_2$};
\node[cf] at (8,-1.3) {\ldots};

\end{tikzpicture}
\caption{Basic structure of a CF-Tree}
\label{fig:cftree}
\end{figure}
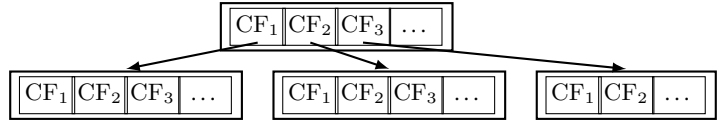

We now briefly introduce the CF-Tree of the improved BETULA version \cite{Lang/Schubert/2020a,Lang/Schubert/2021a},
which improves the numerical accuracy of the original BIRCH CF-Tree~\cite{Zhang/etal/96a,Zhang/etal/97a}.
\index{CF-Tree}

The CF-Tree (Cluster Feature Tree) is a basic height-balanced tree 
storing cluster features (CF). Each BETULA cluster feature \cite{Lang/Schubert/2020a} is a triple
\begin{align}
\CF&:=(n, \mu, \SSE)
\end{align}
where $n$ in this context is the number of data points or their aggregated weight,
$\mu$ denotes the mean vector, and
$\SSE$ is the sum of squared deviations from the mean.
Two BETULA cluster features can be efficiently combined into one:
\begin{align}
\nAB =& n_A + n_B
\label{eq:update-w}
\\
\mu_{AB} =& \mu_A + \tfrac{n_B}{\nAB} (\mu_B-\mu_A)
\label{eq:update-mu}
\\
\SSE_{AB} =& \SSE_A + \SSE_B + n_B (\mu_B-\mu_A)(\mu_B-\mu_{AB})
\enskip. %
\label{eq:update-sse}
\end{align}
A single data point $x$ can be trivially represented by a Cluster Feature $(1,x,0)$.
These update rules also follow from the König-Huygens theorem,
and can be found in \citet{Lang/Schubert/2021a}.
The numerical inaccuracies of the original BIRCH approach were previously observed by \citet{Schubert/Gertz/2018a}.

The CF-Tree is a height-balanced tree: each leaf is a cluster feature that represents data point(s). 
Inner nodes store the aggregated information of their children.
The tree is built by sequentially inserting all data points.
When adding a data point or cluster feature to the tree, it is inserted by traversing the tree and choosing the least distant node on each level.
When a leaf entry is reached the data is added to the leaf entry if the absorption threshold (c.f. Section~\ref{sec:cf-distances}) is not violated.
If the data cannot be added to an existing leaf entry, a new leaf entry is generated.
The threshold can be set based on expert input which results in a tree of variable size but with a fixed accuracy guarantee.
But because we can also add cluster features to the CF-Tree the same way,
we can dynamically rebuild the tree from its leaf entries with an increased threshold 
once a selected maximum number of leaf entries is reached, to reduce the trees memory usage.
In this case the tree is build within a fixed size range but with variable accuracy,
which is beneficial for scenarios where we have memory resource-constraints.

\subsection{Distances for Cluster Features}
\label{sec:cf-distances}

Zhang et al.~\cite{Zhang/etal/96a,Zhang/etal/97a} originally
proposed several distance functions and absorption criteria
for BIRCH cluster features.
Both essentially measure a distance, but distance functions are used to choose insertion subtrees, whereas absorption criteria are used to decide when to add to an existing node, or when create a new node.
It is difficult to argue why the two parameters should ever be set different.\todo{Andreas: neu}
As suggested by \citet{Lang/Schubert/2020a},
we do not distinguish between distances and absorption criteria 
in the following, as there is no benefit of doing so.
\begin{align}
\shortintertext{Euclidean distance:}
\text{D0}(A,B) =& \snorm{\mu_A-\mu_B}_{\phantom{1}} %
\label{eq:D0}
\\
\shortintertext{Manhattan distance:}
\text{D1}(A,B) =& \snorm{\mu_A-\mu_B}_1 %
\label{eq:D1}
\\
\shortintertext{Inter-cluster distance:}
\text{D2}(A,B) =& \sqrt{\tfrac1{n_An_B}\textstyle\sum_{x \in A} \sum_{y \in B}\snorm{x - y}^2}
\label{eq:D2}
\\
\shortintertext{Intra-cluster distance (= diameter absorption criterion):}
\text{D3}(A,B) =& \sqrt{\tfrac1{\nAB(\nAB - 1)}\textstyle\sum_{x, y \in AB} \snorm{x - y}^2}
\label{eq:D3}
\\
\shortintertext{Variance-increase distance:}
\text{D4}(A,B) =& \sqrt{\textstyle
\sum_{x \in AB}\snorm{x{-}\mu_{AB}}^2
{-}\!\sum_{x \in A}\snorm{x{-}\mu_A}^2
{-}\!\sum_{x \in B}\snorm{x{-}\mu_B}^2}
\label{eq:D4}
\\
\shortintertext{Radius absorption criterion:}
\text{R}(A,B) =& \sqrt{\tfrac{1}{\nAB}\textstyle\sum_{x \in AB} \snorm{x - \mu_{AB}}^2}
\label{eq:R}
\end{align}
These distances can be computed efficiently based on the summary statistics
stored in BETULA cluster features.
The corresponding equations and their derivations %
can be found in \citet{Lang/Schubert/2021a}.

\subsection{Hierarchical Clustering with Cluster Features}%
\label{sec:lang-parallels}
While the CF-Tree itself already is a form of hierarchical clustering,
its levels and inner structure are not in a form that is easily interpretable.
Because of this, it is usually only used for data aggregation in preparation of the actual clustering, for which only the leaf entries are used.
Naively, one could just use the centers of the leaf entries and use a standard hierarchical clustering algorithm.
This approach discards the variance information of the clustering features.

The interesting observation now is that linkages and CF distances are not very different. 
We show that there is a correspondence between certain linkages and CF distances,
which can be exploited for clustering by incorporating additional information stored in the cluster features
besides using only the centers.

\begin{table}[tb]\centering
  \caption{Linkage strategy for (squared) Euclidean distances and the corresponding BIRCH distance with their objective function.}
  \label{tab:birch-linkage}
  \begin{tabular}{lll}
  \toprule
  Linkage & Closed form & BIRCH Distance
  \\
  \midrule
  UPGMA & $\tfrac1{n_An_B}\textstyle\sum_{x \in A} \sum_{y \in B}\snorm{x - y}^2 $ &$\text{D2}^2$ 
  \\
  UPGMC & $ \snorm{\mu_A-\mu_B} $ & $\text{D0}^2$ 
  \\
  Ward & $ \tfrac{2n_An_B}{\nAB}\snorm{\mu_A-\mu_B}^2 $& $2\cdot \text{D4}^2$ 
  \\
  \bottomrule
  \end{tabular}
\end{table}

In Tab.~\ref{tab:birch-linkage} we summarize the identified relationships between linkage strategies known from literature and BIRCH distances with their respective object function.
The most obvious similarity can be seen when looking at the Centroid-Euclidean-Distance (D0, Eq.~\ref{eq:D0})
and the Centroid-linkage (Eq.~\ref{eq:centroidlinkage}), which are almost the same. 
The differences between Ward-linkage (Eq.~\ref{eq:ward}) and the Variance-increase-distance (D4, Eq.~\ref{eq:D4}) 
are only in the notation and that D4 squared is Ward, but since BETULA internally uses squared distances
for computational reasons, this difference is trivial. 
The last linkage that can be expressed as a BETULA distance is UPGMA, which is effectively the squared Inter-cluster-distance (D2, 	Eq.~\ref{eq:D2}).
This similarity becomes obvious when replacing the general equation with the one for UPGMA with the squared Euclidean distance:
\begin{align}
d_{\text{UPGMA}}(A,B) =& \tfrac1{n_A \cdot n_B}\textstyle\sum_{a \in A} \sum_{b \in B} d(a,b)\\ 
=& \tfrac1{n_A \cdot n_B}\textstyle\sum_{a \in A} \sum_{b \in B} \snorm{a - b}^2.
\end{align}
While WPGMA and WPGMA cannot have an exact match, we may nevertheless choose D2 respectively D0 as with their ``unweighted''
counterparts because of their close relationship.
With this knowledge we can now meaningfully transition from cluster features into hierarchical clustering
with the Lance-Williams formula by calculating the distance matrix based on the corresponding distances between the cluster features.

We can also do the opposite, and instead of using the classic linkage strategies, we can perform the following
adaptation to hierarchical clustering of cluster features,
while using the distance functions from Section~\ref{sec:cf-distances} instead of a separate linkage strategy.
As in standard hierarchical clustering (e.g., AGNES), we find the smallest non-diagonal value in the distance matrix
to find the best next merge. But instead of combining distances using the Lance-William equation, we can instead
combine the corresponding two cluster features using the update
Eqs.~\eqref{eq:update-w} to \eqref{eq:update-sse}, and compute new distances with respect to the new~CF.

For both cases (using Lance-Williams, and using CF distances),
we can use the approach of Anderberg \cite{Anderberg/73a} and NN-chain \cite{Murtagh/83a}
for acceleration.\index{Nearest-neighbor Chain Algorithm}
While the first does not improve the worst-case complexity of $\O{|\CF|^3}$, it typically performs closer to quadratic in runtime.
The second may yield different results for non-reducible distances (c.f. \cite{Bruynooghe/77a}, Centroid and Median linkage), but guarantees $\O{|\CF|^2}$ runtime;
furthermore, it can be implemented with only linear memory for some linkages.
As the CF-Tree allows us to reduce the data to a constant size 
less than $\O{\sqrt{N}}$ respectively 
$\O{\sqrt[3]{N}}$ cluster features,
we can then perform hierarchical clustering in time linear in the original
data input size~$N$ and within a constant memory limit, making this useful in resource-limited data processing.

\subsection{Experiments}
\label{sec:lang-eval}

We evaluate hierarchical clustering with and without BETULA cluster features.
We are interested in comparing the runtime and quality of aggregated and  non-aggregated algorithms,
but do not compare different linkage strategies.
As baselines, we use the Anderberg~\cite{Anderberg/73a} and NN-Chain~\cite{Murtagh/83a} algorithms
(the latter in an implementation that only uses linear-memory).
For BETULA we allow a maximum of 25\,000 leaf entries, such that no data aggregation takes place for the smallest data sets.
Both of these HAC algorithms can be combined with BETULA in different ways.
We use ``full data'' when not using BETULA aggregation, ``CF~centers'' denotes the naive approach
using the Euclidean distances of the cluster features centers and no weights (found in many implementations of BIRCH).
For ``CF~linkage'', the initial distances are computed using the full cluster feature information,
but afterward the algorithm uses the Lance-Williams equations for hierarchical aggregation.
The ``CF~aggregation'' approach maintains cluster features throughout the hierarchical clustering process.

All algorithms are implemented in the Java framework ELKI~\cite{Schubert/Zimek/2019a}.
By using the same framework for all implementations we try to minimize the effects caused by implementation differences,
as recommended for comparing algorithms~\cite{Kriegel/etal/2017a}.
Each experiment was repeated 10 times with varying input order on a single core of an AMD EPYC\texttrademark~7302 CPU.
Because of our focus on improving the scalability, we may rely on synthetic data for this experiment.
We sample data from both a 5-dimensional uniformly distributed hypercube,
respectively from a combination of 500 5-dimensional Gaussian clusters.
While the uniform distribution is supposed to adversely affect the aggregation quality of BETULA,
the Gaussian clusters are well-suited for this type of aggregation.

First we look at the runtime analysis with 50\,000 data points, the biggest data set the
baseline Anderberg implementation can process.\footnote{Because the array size reaches the $2^{31}$ array length limit of Java.}
As Tab.~\ref{tab:lang-rt-comparison} shows, the NN-Chain algorithm can be significantly faster than the Anderberg algorithm
(at least for this low-dimensional data set).
While we still largely limit the data aggregation of BETULA (set to a maximum of 25\,000 leaves),
the number of CFs obviously is the main contributor to runtime as seen when comparing the
results on uniform data with those on Gaussians. The design of BETULA does not allow for
an exact control of this number, but when the given maximum is reached a smaller tree is
built from the current leaves. By choosing a smaller limit, an even larger speedup over the
baseline algorithms would be possible.
Because the data aggregation performed by BETULA is deterministic, the same input leads to the same tree (and hence number of CF in Tab.~\ref{tab:lang-rt-comparison}) independent of the clustering step used afterwards.

\begin{table}[bt!]\centering
  \caption{Average runtime in seconds and number of cluster features after BETULA initialization for different algorithms and data generators with $N$\,=\,50\,000 and $d$\,=\,5 dimensions.}
  \label{tab:lang-rt-comparison}
  \setlength{\tabcolsep}{4pt}\small\renewcommand{\arraystretch}{1.2}%
  \begin{tabular}{l|l|rr|rr}
  \toprule
    \multirow{2}{*}{Algorithm} & \multirow{2}{*}{Input} & \multicolumn{2}{c|}{Uniform} & \multicolumn{2}{c}{Gaussian} \\
    & & Runtime & |CF| & Runtime & |CF| \\
  \midrule
    Anderberg & full data & 142.96 & - & 147.05 & - \\
    Anderberg & CF centers & 12.99 & 17482.4 & 4.88 & 9496.7 \\
    Anderberg & CF linkage & 12.85 & 17482.4 & 4.71 & 9496.7 \\
    NN-Chain & full data & 49.64 & - & 49.44 & - \\
    NN-Chain & CF aggregation & 11.39 & 17482.4 & 4.36 & 9496.7 \\
    NN-Chain & CF linkage & 7.18 & 17482.4 & 3.15 & 9496.7 \\
  \bottomrule
  \end{tabular}
\end{table}

\begin{figure}[tb!]
  \captionsetup[subfigure]{justification=centering}
  \begin{subfigure}{.48\linewidth}\centering
  \includegraphics[width=\linewidth]{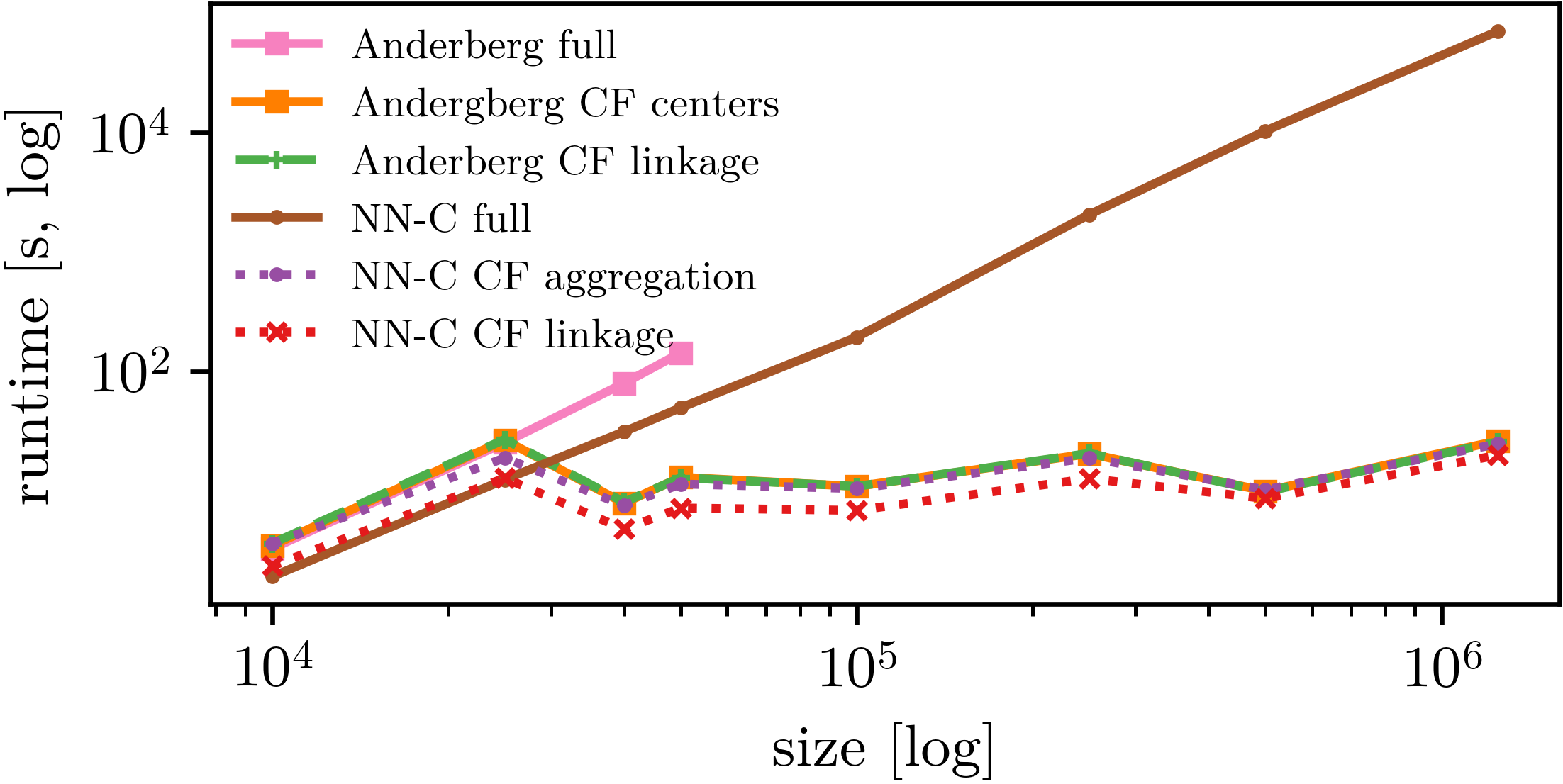}
  \caption{UPGMC}
  \label{fig:lang-rt-a}
  \end{subfigure}%
  \hfill%
  \begin{subfigure}{.48\linewidth}\centering
  \includegraphics[width=\linewidth]{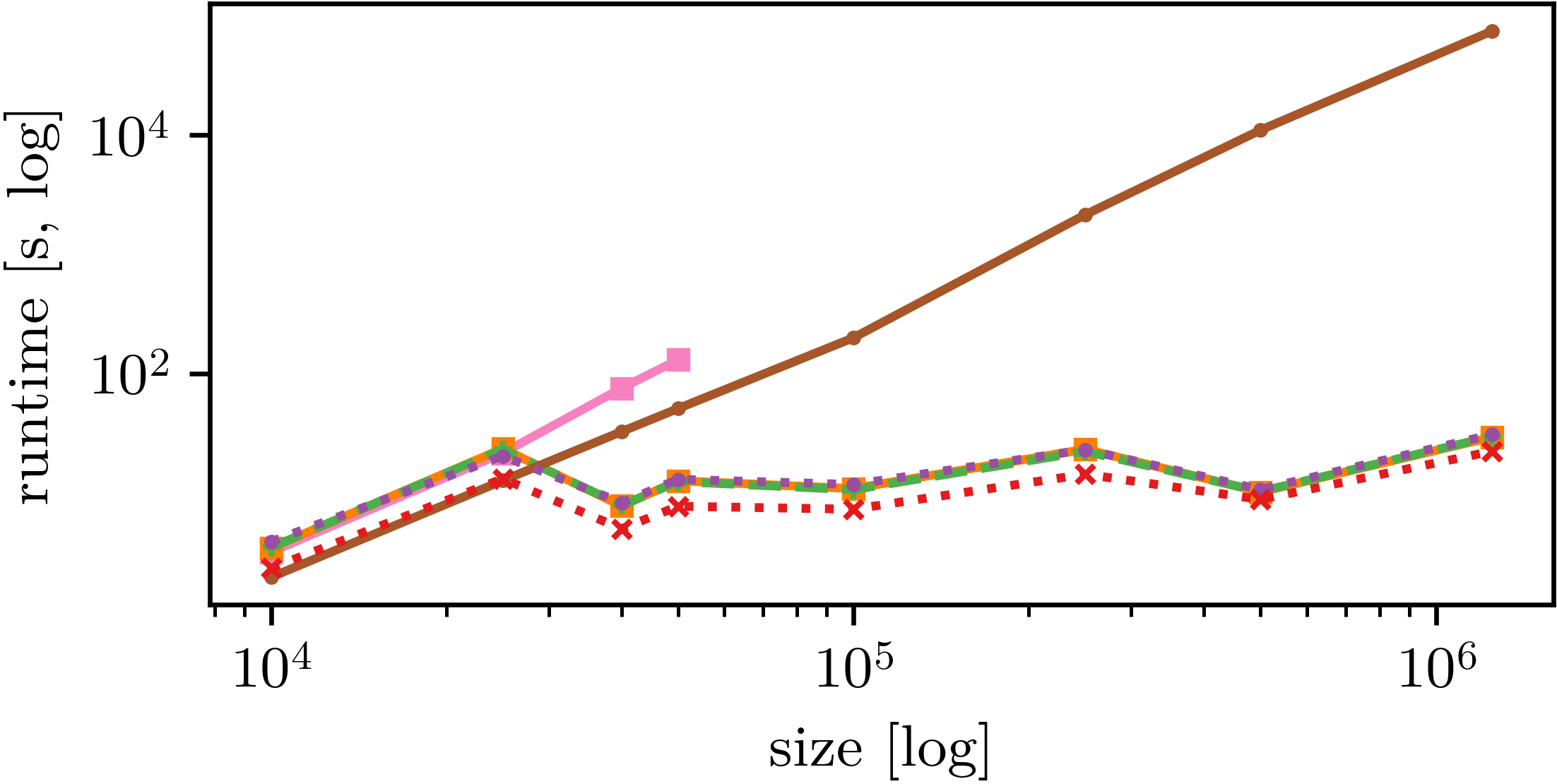}
  \caption{Ward}
  \label{fig:lang-rt-b}
  \end{subfigure}
  \caption{Runtime versus data set size on uniformly distributed data.}
  \label{fig:lang-rt}
\end{figure}

Next we look at the scalability of our approach. 
Fig.~\ref{fig:lang-rt} shows the runtime of the algorithms for Centroid (UPGMC) and Ward linkage on various data set sizes
in a log-log-plot.
We can see the quadratic increase in runtime of the baseline NN-Chain and Anderberg algorithms.
For the Anderberg baseline, only times up to 50\,000 points are given because of Java array size restrictions,
but scaling would be at least as bad as for the NN-Chain algorithm.
The runtime for all variants that use BETULA for data aggregation seems to fluctuate around some constant value.
This is caused by changes in the number of tree leaf CFs (because the results are averaged over multiple
permutations of the data set, tree sizes and tree rebuilds are not constant for a particular data set size).
Even for big data set sizes, the CF-tree construction phase which has a runtime in~$\O{N}$ plays a minor role
compared to the later hierarchical clustering phase with its $\O{|\CF|^2}$ runtime; and reading the input data once is unavoidable in most applications.

\begin{table}[tb!]\centering
  \caption{Root mean squared deviation for different linkages, algorithms, and data sets with {$N$\,=\,50\,000}. All values are given as mean value plus minus standard deviation over 10 runs.}
  \label{tab:lang-rmsd-comparison}
  \setlength{\tabcolsep}{.8em}\small\renewcommand{\arraystretch}{1.2}%
  \begin{tabular}{lllrrr}
  \toprule
    Algorithm & Input & Generation & UPGMC & UPGMA & Ward \\
  \midrule
    Anderberg & full data & Uniform & 10.34 $\pm$ 0.00 & 10.11 $\pm$ 0.00 & 10.12 $\pm$ 0.00 \\
    Anderberg & CF linkage & Uniform & 10.53 $\pm$ 0.03 & 10.29 $\pm$ 0.02 & 10.27 $\pm$ 0.01 \\
    NN-Chain & full data & Uniform & 10.17 $\pm$ 0.04 & - & 10.17 $\pm$ 0.04 \\
    NN-Chain & CF linkage & Uniform & 10.36 $\pm$ 0.03 & - & 10.35 $\pm$ 0.02 \\
  \midrule
    Anderberg & full data & Gaussian & 3.56 $\pm$ 0.00 & 3.51 $\pm$ 0.00 & 3.38 $\pm$ 0.00 \\
    Anderberg & CF linkage & Gaussian & 3.56 $\pm$ 0.00 & 3.53 $\pm$ 0.03 & 3.40 $\pm$ 0.01 \\
    NN-Chain & full data & Gaussian & 5.25 $\pm$ 0.05 & - & 5.25 $\pm$ 0.05 \\
    NN-Chain & CF linkage & Gaussian & 5.09 $\pm$ 0.05 & - & 4.95 $\pm$ 0.08 \\
  \bottomrule
  \end{tabular}
\end{table}

The quality of a hierarchical clustering is hard to evaluate properly,
because it very much depends on the data set and application.
A thorough evaluation of a clustering on real data will usually require manual inspection by a domain expert.
For our experiments, we chose to simply compare the variability of the clusters when cut into $500$ clusters,
assuming that a result with less spread also indicates a better clustering.
Tab.~\ref{tab:lang-rmsd-comparison} shows the root mean squared deviation~(RMSD) for all relevant algorithms for the data sets with 50\,000 data points.
Here, the runtime improvements with BETULA were significant but the difference in quality for all algorithms
is very small for the uniform data set (within the variability caused by NN-Chain using a different processing order than
Anderberg).
The results on the Gaussian data warrants further discussion.
On this data set, which is favorable to the assumptions of BETULA, the negative effect of the data aggregation when combined with Anderberg is even smaller.
There is no measurable difference for UPGMC and only slightly worse results for UPGMA and WARD.
The NN-Chain algorithm on the other hand suffers from its known differences to the Anderberg algorithm
(making greedy locally optimal choices, as opposed to choosing the global optimum).

\begin{figure}[tb!]
  \captionsetup[subfigure]{justification=centering}
  \begin{subfigure}{.48\linewidth}\centering
  \includegraphics[width=\linewidth]{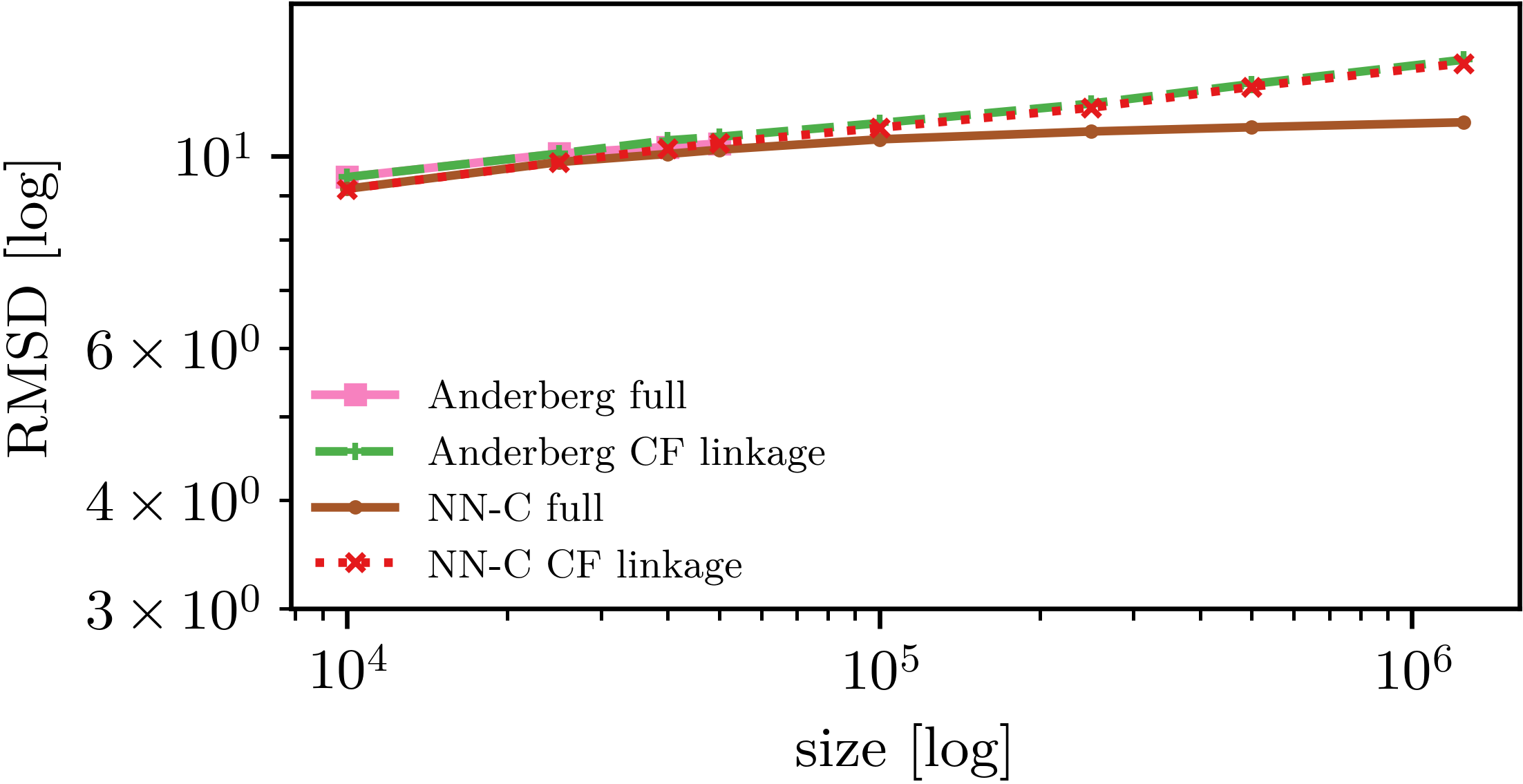}
  \caption{Uniform}
  \label{fig:lang-q-a}
  \end{subfigure}%
  \hfill%
  \begin{subfigure}{.48\linewidth}\centering
  \includegraphics[width=\linewidth]{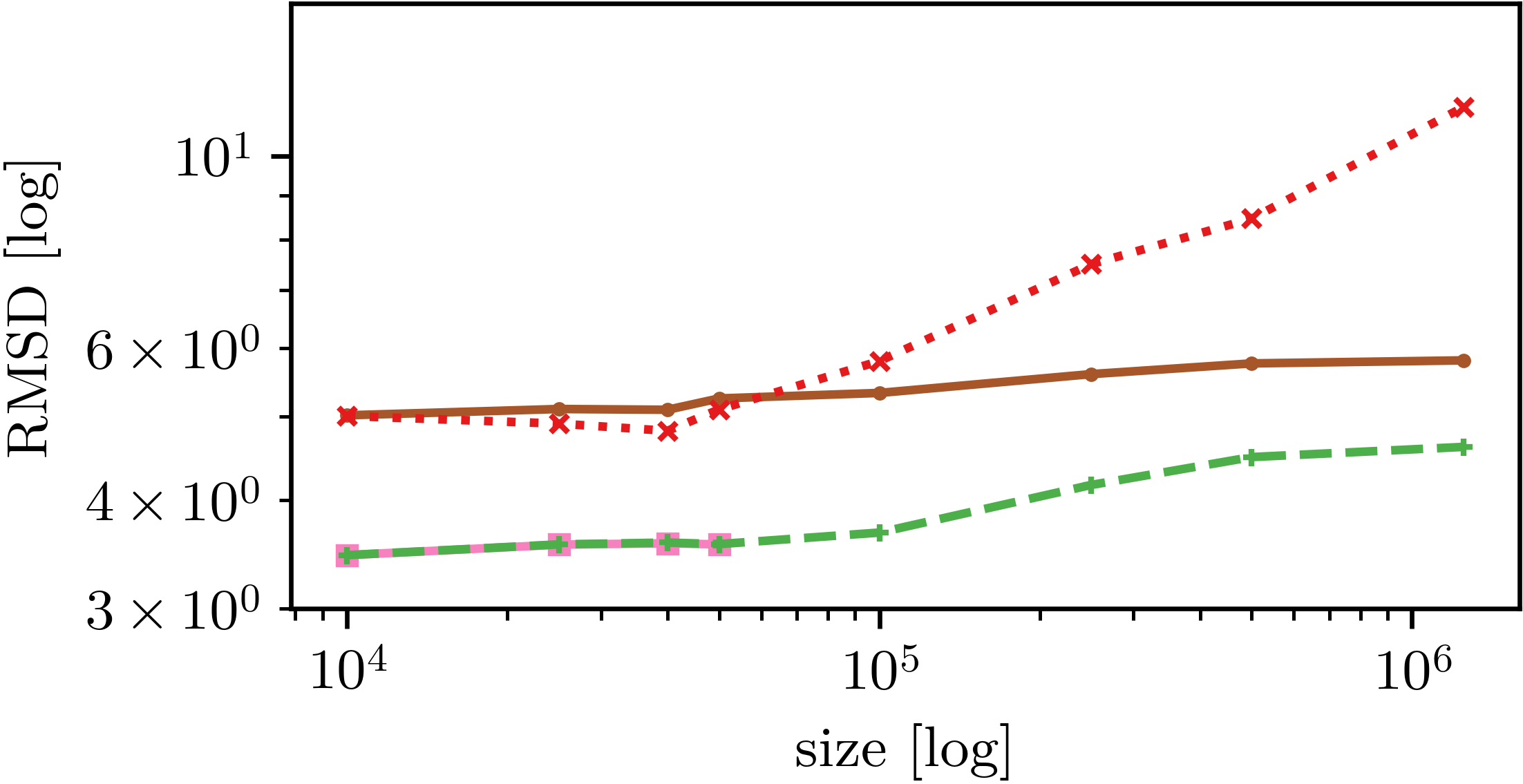}
  \caption{Gaussian}
  \label{fig:lang-q-b}
  \end{subfigure}
  \caption{Root mean squared deviation %
  versus data set size using centroid (UPGMC) linkage for both data set generators and $k$\,=\,500.}
  \label{fig:lang-q}
  \end{figure}

Finally we evaluate the scalability of our approach.
Fig.~\ref{fig:lang-q} shows the root mean squared deviation of the $k$\,=\,500 clusters.
For $N$ = 25\,000, where no aggregation takes place, the results are the same, with and without BETULA.   
The results for the data sets with more entries are similar. while the number of cluster features used stays below 25\,000, the quality only is impacted slightly.
On the uniform data the difference in quality between the algorithms is small.
The Gaussian data shows that the difference between the NN-Chain and Anderberg algorithms is bigger than that of the data aggregation with BETULA.
The only outlier is the combination of BETULA and NN-Chain, which shows a noticeablly worse result.

\subsection{Conclusion}
In this section we discussed how the scalability of hierarchical clustering can be improved
by integrating data aggregation techniques from BIRCH (or its more stable variant BETULA).
We show how hierarchical linkages relate to particular BIRCH distance criteria,
and that some criteria improve the clustering for the same metric. 
We use this relation to accelerate the hierarchical clustering with small effects on the
quality of the clustering while keeping most benefits of hierarchical approaches
and expanding it to data set sizes not practical for the standard approaches. 
This optimization allows the usage of hierarchical clustering on small or embedded systems
with limited memory by using data aggregation to decouple the total data size from the
input data size of the much more expensive hierarchical clustering step,
leading to better scalability. While there is some loss in clustering quality,
it is small enough for most use cases of explorative data analysis, i.e.,
we will still be able to make meaningful choices for the subsequent steps in our data analysis process.

\fi

\let\O\undefined%

\egroup

\printbibliography
\end{document}